\documentclass{Interspeech2024}

\title{LLaMA based Punctuation Restoration With Forward Pass Only Decoding}

\name[affiliation={1}]{Yutong}{Pang}
\name[affiliation={1}]{Debjyoti}{Paul}
\name[affiliation={1}]{Kevin}{Jiang}
\name[affiliation={1}]{Xuedong}{Zhang}
\name[affiliation={1}]{Xin}{Lei}

\address{
  $^1$Meta, USA}
\email{yutongpang@meta.com, debjyotipaul@meta.com, 
kevinjiang@meta.com,  xuedongzhang@meta.com, leixin@meta.com}

\renewcommand{\headrulewidth}{0pt}
\renewcommand{\footrulewidth}{0pt}


\usepackage[dvipsnames]{xcolor}
\usepackage{multirow}
\definecolor{DarkGreen}{HTML}{006608}
\newcommand{\itbf}[1]{\textit{\textbf{#1}}}

\algnewcommand{\LeftComment}[1]{\Statex {\color{DarkGreen} \(\triangleright\)  #1}}
\algnewcommand{\LeftCommentSpace}[1]{\Statex {\color{DarkGreen} \hspace{9mm}\(\triangleright\)  #1}}

\begin{document}

\maketitle

\begin{abstract}
This paper introduces two advancements in the field of Large Language Model Annotation with a focus on punctuation restoration tasks. Our first contribution is the application of LLaMA for punctuation restoration, which demonstrates superior performance compared to the established benchmark. 

Despite its impressive quality, LLaMA faces challenges regarding inference speed and hallucinations. To address this, our second contribution presents Forward Pass Only Decoding (FPOD), a novel decoding approach for annotation tasks. This innovative method results in a substantial 19.8x improvement in inference speed, effectively addressing a critical bottleneck and enhancing the practical utility of LLaMA for large-scale data annotation tasks without hallucinations.

The combination of these contributions not only solidifies LLaMA as a powerful tool for punctuation restoration but also highlights FPOD as a crucial strategy for overcoming speed constraints. 
\end{abstract}

\section{Introduction}
Automatic Speech Recognition (ASR) plays a vital role in numerous domains involving human-computer interaction \cite{hinton2012deep} \cite{mohamed2011acoustic} \cite{sak2014long}. However, the outputs of many ASR systems often lack punctuation. Punctuation restoration in the context of ASR output is a crucial component \cite{tilk2016bidirectional} \cite{courtland2020efficient}, essential for enhancing the overall utility, user experience, and comprehensibility of transcribed speech. Restoring the punctuation will make the raw ASR output more coherent, with improved intractability.

The field of punctuation restoration encompasses two distinct techniques: cascade methods, exemplified by models like BERT \cite{devlin2018bert}, commonly applied independently to Automatic Speech Recognition (ASR) outputs in spoken domains without punctuation \cite{puaics2022capitalization}. These cascade models function as standalone systems, addressing the punctuation restoration task sequentially. On the other hand, the End-to-End (E2E) approach, represented by models such as Recurrent Neural Network Transducer (RNNT) or Whisper \cite{radford2023robust}, trained in an end2end fashion, incorporates built-in punctuation output. This category of techniques streamlines the punctuation restoration process. However, both approaches face challenges, the former requiring independent but domain-aligned training data and evaluation effort, and the latter compels to use large amounts of high-quality supervised data containing punctuation paired with audio, which is a bottleneck for scaling ASR systems to new domains and languages requiring punctuation restoration.

Recognizing the significance of the punctuation restoration task and the challenges posed by previous models, our work introduces an approach that leverages the capabilities of LLaMA. Acknowledged for its effectiveness in various language-related tasks, LLaMA emerges as a compelling alternative that surpasses existing benchmarks across numerous Natural Language Processing (NLP) tasks \cite{touvron2023llama} \cite{touvron2023llama2}. Additionally, with LoRA fine-tuning \cite{hu2021lora}, which demands significantly less supervised training data, we achieve comparable and even superior performance for punctuation restoration compared to traditional methods introduced previously. This approach addresses both the quality effectiveness and scale-up concerns associated with punctuation restoration in diverse languages and domains.

In our exploration of LLaMA-based punctuation restoration, we present a range of strategies. Initially, we delve into the traditional approach with auto-regressive generation. Subsequently, we explore techniques to address the inherent challenge of inference speed in LLaMA. The first of these strategies involves speculative decoding, showcasing improvements in inference speed while maintaining the quality of generated outputs exactly the same as the original base model. Finally, we present a new forward pass only approach, eliminating the need for auto-regressive generation entirely. This novel approach results in a substantial boost in inference speed.

Our contribution not only establishes LLaMA as a potent alternative for achieving high-quality punctuation restoration but also introduces practical enhancements to overcome the challenges associated with inference speed.
\section{Proposed Method}
In this section, we described the proposed forward pass only method to restore the punctuation. At the same time, we will compare it with other decoding methods: the auto regressive decoding and speculative decoding.
\subsection{Auto Regressive Generation}
Auto-regressive generation refers to a process in which a language model generates output (e.g., text) one token at a time in a sequential manner. At each step, the model predicts the next token based on the context of the preceding tokens it has generated. This process is "auto-regressive" because the model's own outputs become part of the input for predicting subsequent tokens. Inference from LLaMA auto-regressively is slow - decoding K tokens takes K sequential run of the model. 
\subsection{Speculative Decoding}
We have already seen that the auto regressive generation is a very slow generation process. Speculative Decoding is introduced to improve it \cite{leviathan2023fast}. It refers to the process of using an assistant model to help the decoding process to prevent going through auto regressive decoding for most cases. Here is a brief explanation of how speculative decoding works:
\begin{itemize}
\item We first use the assistant model (usually a small distilled student model) to generate the output auto regressively
\item Then we send the output to the large main model (usually a large teacher model), and only perform verification forward passes
\item If the verification is successful (the main model agrees with the assistant model), then we directly use the assistant model output as final output. Otherwise, we need to run the full auto-regressive generation with the large main model to get a “better” output.
\item Since for the cases with successfully verified results, we only run the auto-regressive generation with the fast assistant model and only perform verification forward pass with the slow main model, the decoding process is sped up substantially.
\end{itemize}
Speculative decoding could help us to improve the inference speed; however, we still need to train the distilled student model, and auto-regressive generation is still needed for all the student model pass and some of the base model pass (the case failed with forward verification).
The inference speed limit is totally dependent on the quality and size of the student model. And the general inference speed improvement is usually less than 2X \cite{chen2023accelerating}.

\begin{figure}[!b]
  \centering
  \vspace{-3mm}
  \includegraphics[width=\linewidth]{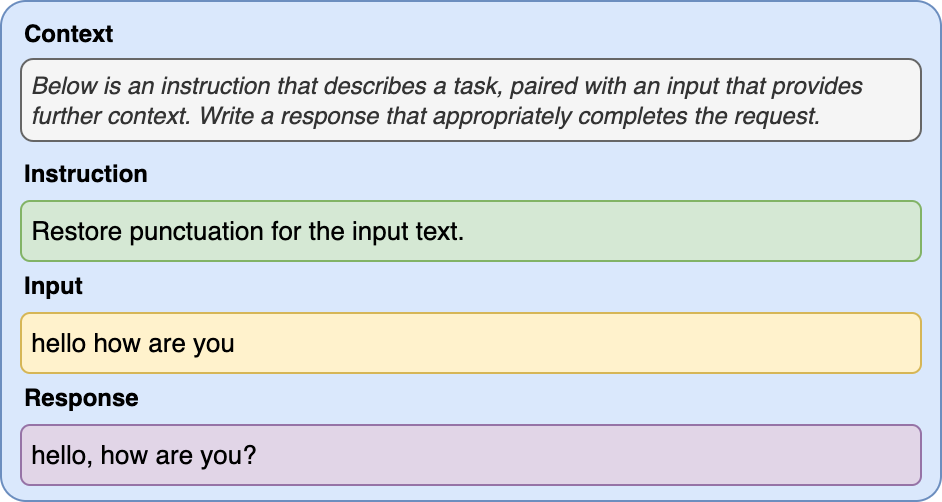}
  \caption{LoRA based Llama2 finetuning prompt template with example instruction, input and response.}
  \label{fig:lora_template}
\end{figure}

\subsection{Forward Pass Only Decoding}
\begin{figure}[t]
  \centering
  \includegraphics[width=\linewidth]{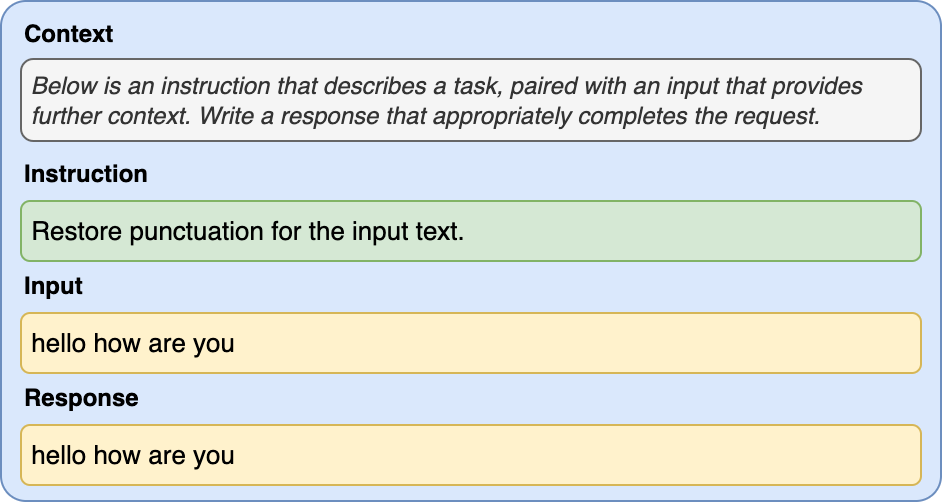}  
  \caption{Directly feeding input as response in prompt for forward pass only decoding (FPOD) scheme.}
  \label{fig:forward_decoding_2}
  \vspace{2mm}
\end{figure}

\begin{figure}[!t]
  \centering
  \includegraphics[width=0.95\linewidth]{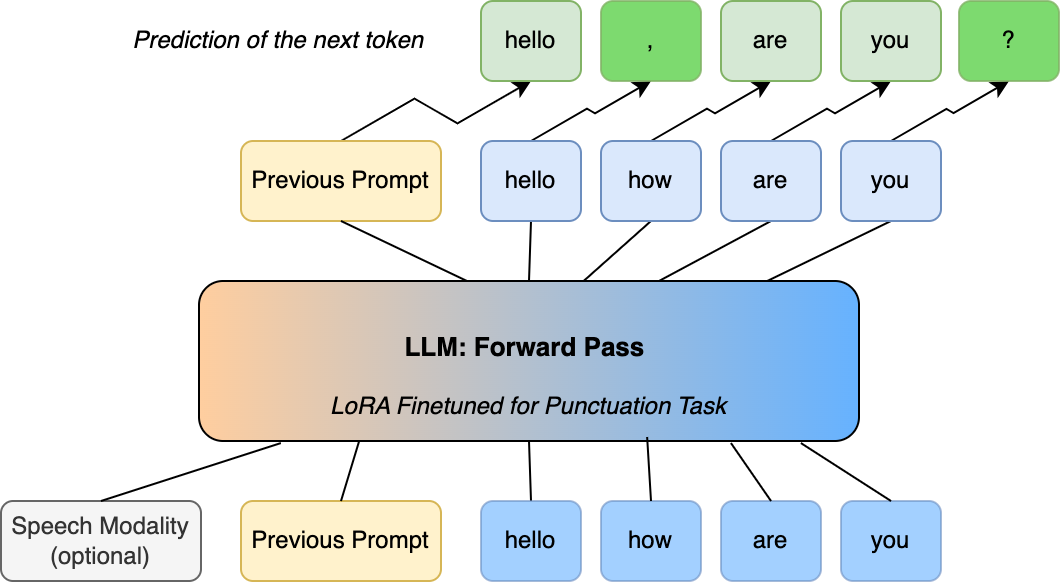}
  \caption{FPOD for punctuation restoration}
  \label{fig:forward_decoding_3}
  \vspace{-4mm}
\end{figure}

\itbf{Concept.} In this section, we introduce forward pass only decoding (FPOD), which can totally discard the auto-regressive generation step. 
We can employ FPOD for tagging, edit, prepend, and append-based text-enhancement tasks.
For ease of understanding, we will proceed with the punctuation restoration task as an example for the decoding scheme. The following explains the step-by-step procedure to achieve the punctuation restoration task with FPOD. Also, the detailed algorithm is illustrated in Algorithm~\ref{alg:forward_punctuation}. 
\begin{itemize}
\item We will first use LoRA fine tuning to finetune the LLaMA2 model for punctuation restoration task following the prompt template in Figure~\ref{fig:lora_template}. 
\item We directly feed the following prompt for forward pass; notice the input is copied for the response as shown in Figure~\ref{fig:forward_decoding_2}.
\item In a single forward pass, for each token in the response, we obtain a prediction of the next token, as shown in Figure~\ref{fig:forward_decoding_3}. 
If the prediction of the next token is a punctuation symbol, we will prepend it to the current token.
\end{itemize}
\begin{algorithm}[!b]
\caption{ForwardPassOnlyDecoding($M$,  $x_{[1:T+L]}$)}\label{alg:forward_punctuation}
\begin{algorithmic}
\Require $M$, $x_{[1:T+L]}$
\LeftComment{$M$ is a task-based finetuned LLaMA2 model with LoRA,}
\LeftComment{$x_{[1:T+L]}$ input prompt tokens, $T$ is length of prompt template length, T+L is the total prompt length including the response part.}
\Ensure {\it resText}
\LeftComment{\text{Forward} pass in parallel to get next token predictions, so we convert a generation task into a verification task}
\State $y_{[1:T+L+1]} \leftarrow M(x_{[1:T+L]})$   
\LeftComment{\text{Append Symbol} $i$ restoration interation.}
\For{$i$ $\leftarrow$ $T+1$ \text{to} $T+L+1$ } \Comment{{\color{DarkGreen} In the response region.}}
    \If {$y_i$ is {\it digit} and $x_i$ is {\it space}} 
        \State  {\it continue} \Comment{ {\color{DarkGreen} delete current space}}
    \EndIf
    \If{$y_i$ is {\it appendToken}} \Comment{ {\color{DarkGreen} Append Symbol found}}        
        \LeftCommentSpace{{\color{DarkGreen}Append symbol}}
        \If{$i$ = $T+L+1$} \Comment{{\color{DarkGreen} for last token}}
            \State \text{\it resText} $\leftarrow \text{\it resText} + y_{i}$
        \Else \Comment{{\color{DarkGreen} for non-last token}}
            \State \text{\it resText} $\leftarrow \text{\it resText} + {y_i}+ {x_i}$
        \EndIf
    \Else
        \State \text{\it resText} $\leftarrow \text{\it resText} + x_{i}$
    \EndIf\
\EndFor \\
\Return {\it resText}
\end{algorithmic}
\end{algorithm}


By employing this method, we can effectively restore punctuation, resulting in sentences like "hello, how are you?". This approach converts the original generation task into a verification task, significantly enhances the inference speed of punctuation restoration compared to traditional auto-regressive methods. Furthermore, we utilize the frozen LoRA fine-tuned model, eliminating the need for additional training such as token classification for punctuation task.\vspace{2mm}

In addition to enhancing speed, the use of forward pass only decoding ensures that the token lengths and the original sentence structure (with punctuation modifications only) remain unaltered. This method effectively mitigates the issue of hallucination, a common problem associated with the auto-regressive approach.\\
\itbf{Limitations.}
Decoding solely through the forward pass appears highly efficient and straightforward; nevertheless, certain details require careful consideration:
\begin{itemize}
\item In general, the performance of the large language model (LLM) is usually worse for ``super long" input context, which is often more important for punctuation restoration.
\item With forward only decoding, the given token prediction only depends on the previous token history. So let’s see if the history is ``hello how are you”, and we want to predict the next token of you, ideally should be ``?”. However, because the previous history does not contain any punctuation, the model behavior may be different from the auto-regressive generation process.
\end{itemize}\vspace{2mm}\
\itbf{Solutions.}
To address the first limitation i.e., decoding longer text, we use a simple sliding window with padding approach, illustrated by the following Figure~\ref{fig:sliding_window}.
To solve the second limitations i.e., context dependant decoding, instead of one pass forward decoding, we will split the process into the following step as {\it Recursive FPOD}:
\begin{figure}[t]
  \centering
  \hspace{-5mm}
  \includegraphics[width=1.05\linewidth]{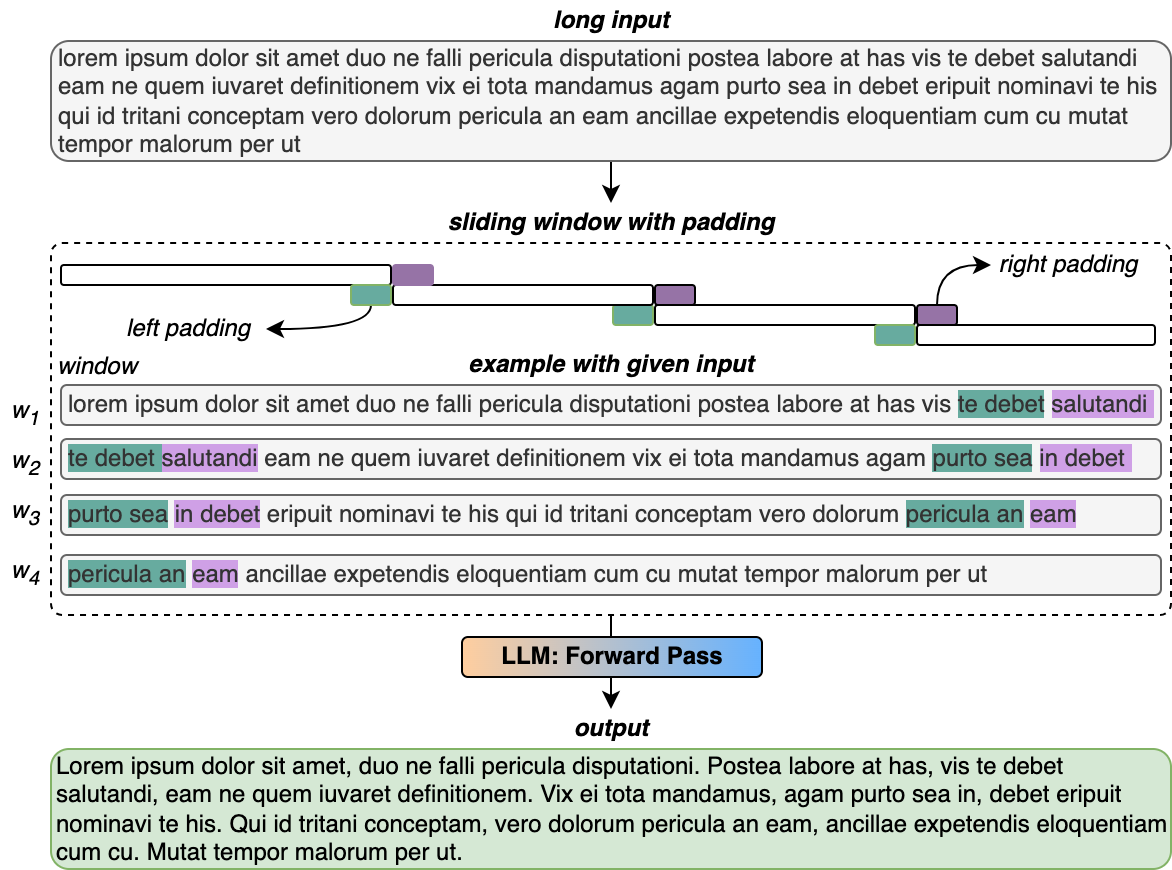}
  \caption{Sliding window with padding approach for long input text.}
  \label{fig:sliding_window}
\end{figure}
\begin{itemize}
\item Iterate through the input tokens “hello how are you”, we will update the sentence once we find a punctuation prediction.
\item In this case, “hello how are you” $\rightarrow$ “hello, how are you” $\rightarrow$ “hello, how are you?”
\item So instead of one pass forward decoding, we pass the input sentence two times to the forward decoding process.
\end{itemize}
% \blue{TODO: (a) Discussion on punctuation frequency in English corpus. (b) average recursive decoder call analysis based on the expectation of punctuation in a language. https://arxiv.org/pdf/2211.17192.pdf}\\
\itbf{Improvement Factor.} 
Lets analyze the improvement factor running recursive FPOD with respect to auto-regressive generation. \\
The probability of accepting output from forward pass $y_{[1:T+L]} \leftarrow M(x_{[1:T+L]})$ as it is, from Algorithm \ref{alg:forward_punctuation}, i.e., without modification of token for punctuation task be $\beta$. Then $E(\beta)$ is the measure of how efficient FPOD predicts with respect to regressive generation. Now, simplifying the assumption that the 
$\beta$s are independent and identically distributed (i.i.d.), the expected number of tokens is not followed by punctuations when recursive FPOD is not needed, and the forward pass result is readily accepted. We  dub it as \textit{acceptance rate}
and denote as $\alpha = E(\beta)$. Then the expected number of
tokens produced by a single run of recursive FPOD on the length of tokens $L$ is a capped
geometric variable series is \\
$$E(\#token) = \frac{1-\alpha^L}{1-\alpha}$$. 

% \red{
% The expected #token seems right, however, when we first meet a punctuation, we need to recursive apply this again, so make it a little bit tricky here, let's E(#token) is the parallel forward verification pass, which will take $t_1$, and the normal auto-regressive will take $t_2$, then $t_2/t_1 =\sigma E(#token)$, where $\sigma$ is slightly less than 1, then we need to calculate the E(#token) again for the second recursive call, and add them all together, which is a little bit tricky here 
% }\\
% \blue{The formula $E(\#token)$ is calculated keeping recursive FPOD in mind. See the Sec 3.3 in reference paper (https://arxiv.org/pdf/2211.17192.pdf). It says, ``with the i.i.d. assumption our algorithm reduces the number of calls to the target model by a factor" $\frac{1-\alpha^L}{1-\alpha}$.\\
% I agree with your parallel Forward pass vs regressive step improvement factor. The factor $t2/t1$ or $\sigma$ could be made into a complex calculation. However, lets assume it as a time-efficiency factor $\eta$, as addressed in the next paragraph, as a wholly simplified factor for time-efficiency for running forward pass vs regressive generation for a fixed length L tokens.}

 The above factor is similar to speculative decoding expected token generation \cite{leviathan2023fast}. Moreover, for Algorithm~\ref{alg:forward_punctuation} we need to consider a time efficiency factor for running forward pass decoding w.r.t. regressive generation. We introduce $\eta$, a time-efficiency factor to attribute running one step of forward pass decoding vs. one step regressive generation, where $\eta \leq 1$ but very close to $\approx1$, in the later experiment section, we will give an estimate of the $\eta$. Since forward pass predicts for $L$ tokens in parallel with multiprocessing, ideally taking the same time as one token prediction with regressive generation, with slight overhead for multiprocessing. Then, the overall expected improvement factor in token generation is
 $$\text{\it Improvement Factor (IF)} = \eta\frac{1-\alpha^L}{1-\alpha}$$

% Let $\alpha$, the expected number of tokens not followed by punctuation, that is when recursive FPOD is not needed and the forward pass result is readily accepted. We can also dub this term as \textit{acceptance rate}. The expected number of tokens generated by recursive FPOD in comparison to auto-regressive will be bound by the geometric progression $\frac{1-\alpha^L}{1-\alpha}$ times in the same number of iterations, where $L$ is the response length \cite{leviathan2023fast}. Furthermore, let $\eta>1$, the time-efficient factor for running forward pass in comparison to auto-regressive generation scheme over $L$ tokens. The overall improvement factor is $\eta\frac{1-\alpha^L}{1-\alpha}$.

Let's conduct an empirical analysis to gauge the enhancement factor of FPOD for the punctuation task. Drawing from a frequency distribution analysis of punctuation marks in English across extensive corpora \cite{sun2019frequency}, we can approximate around 91,000 punctuation marks (including commas, periods, and question marks) per million words, equating to roughly 9 punctuation marks per 100 words. We can reasonably assume an average number of tokens per word for LLaMA models, denoted as $\kappa$, where $\kappa \geq 1$. This implies we expect to encounter approximately 9 punctuation marks every 100$\kappa$ tokens. However, for the sake of simplicity, we'll consider $\kappa=1$.
Hence, $\alpha$ is, 
\[
\alpha= 1- \frac{9}{100} = 0.91 \tag*{let $\kappa=1$}
\]
Therefore, the improvement factor for the punctuation task is
\[
\text{\it IF} = \eta\frac{1-\alpha^{L}}{1-\alpha}=\eta\frac{1-0.91^{50}}{1-0.91}\approx11\eta \tag*{for  $L=50$}
\]\vspace{1mm}\\
\itbf{Applications.} As mentioned earlier, FPOD can be promising in various applications such as tagging, verification, and text enhancement. For instance, it can be utilized for tasks like entity tagging, verifying speech recognition transcriptions for quality control, and text normalization or inverse-text normalization.

\section{Experiments}
In this section, we verify the effectiveness of the proposed forward pass only decoding method for punctuation restoration. We will compare the F1 score as a metric for punctuation restoration quality \cite{puaics2022capitalization}. In all the experiments, the punctuation restoration is applied directly to ASR output reference (without punctuation). We will also compare the inference speed, measured by tokens/second for each decoding method. The detailed setup and results of each experiment are also described.
\subsection{LoRA Finetuned Model for Punctuation Restoration}
The punctuation restoration model is trained with Lora Fine tuning on the 13B LLaMA2 model, and the training data is 20k of train-clean-360 data from Librispeech-PC dataset \cite{meister2023librispeech}. The prompt template for LoRA fine-tuning is described in Figure~\ref{fig:lora_template}. After the LoRA fine-tuning process, we can get the merged model for the punctuation restoration task. We run the knowledge distillation (KD) with the same training dataset to get the distilled assistant model (350MB) \cite{liu2024mobilellm}. 
\subsection{Punctuation Benchmark with Different Decoding Methods}
In this experiment, we will evaluate and compare the F1 scores and inference speeds for each decoding method. We used the Librispeech-PC dataset test split for benchmark \cite{meister2023librispeech}. The results indicate that speculative decoding (SD) provides a 1.95x improvement in inference speed, maintaining the same F1 score as auto-regressive (AR) generation with the 13B parameters model. The forward pass only decoding (FPOD) method yields a remarkable 19.84x boost in inference speed, with only a minor decrease in the F1 score for commas. These findings suggest that the forward pass only method is a compelling option for large-scale punctuation restoration task.

\begin{table}[h]
  \caption{Punctuation Benchmark Result (Librispeech-PC) with Different Decoding Method}
  \label{tab:experiment_1}
  \centering
  \begin{tabular}{ l|ccc|r }
    \toprule
    ~ & \multicolumn{3}{c|}{\bf F1 Scores}  & ~ \\
    \textbf{Method} & \textbf{`?'} & \textbf{`.'} & \textbf{`,'} & \textbf{tokens/s} \\
    \midrule
    Auto Regressive (AR) & 0.80 & 0.84 & 0.74 & 88.72 \\
    Speculative (SD) & 0.80 & 0.84 & 0.74 & 173.43 \\
    LLaMA2 (FPOD) & 0.79 & 0.85 & 0.66 & {\bf 1760.30} \\ 
    \bottomrule
  \end{tabular}
\end{table}

\begin{table}[b]
  \caption{LLaMA based annotation vs. RNNT and Whisper Model on Video ASR eval}
  \label{tab:experiment_2}
  \centering
  \begin{tabular}{ l|ccc }
    \toprule
    ~ & \multicolumn{3}{c}{\bf F1 Scores}\\
    % \textbf{Method} & \textbf{`?'} & \textbf{`.'} & \textbf{`,'}\\
    \textbf{Model} & \textbf{`?'} & \textbf{`.'} & \textbf{`,'} \\
    \midrule
    RNN-T 1B params & 0.63 & 0.68 & 0.40  \\
    Whisper Large v2 & 0.58 & 0.65 & 0.46  \\
    LLaMA2 (FPOD) & 0.79 & 0.82 & 0.63  \\
    LLaMA2 (recursive FPOD) & \textbf{0.93} & \textbf{0.95} & \textbf{0.87}  \\
    \bottomrule
  \end{tabular}
  \vspace{3mm}
\end{table}

\subsection{LLaMA based model vs. RNNT model and Whisper for long input utterance}
In this section, we aim to compare and assess the performance of the LLaMA-based punctuation restoration model against the RNNT and Whisper models, as discussed in the preceding section. Our focus is primarily on long input utterances within the video ASR domain. We utilize an in-house, human-annotated dataset for evaluation, comprising 1.2k long utterances, each averaging 5 minutes. The reference for the evaluation set includes 808 question marks (?), 9837 periods (.), and 6457 commas (,). We employ a recursive forward decoding technique with a window for this study. As shown in Table~\ref{tab:experiment_2}, the results suggest that the LLaMA2-based model, both with FPOD and recursive FPOD method, can improve the F1 score for all punctuation marks. These improvements surpass the performance of both the RNNT and Whisper models. Notably, the utilization of recursive FPOD further amplifies the F1 score by a substantial margin.
Regarding inference speed, the adoption of recursive FPOD achieves an impressive rate of 959.1 tokens/s for long input texts. This represents a remarkable 10.8x improvement compared to the auto-regressive baseline of 88.72 tokens/s, as demonstrated in Table~\ref{tab:experiment_1}. Here we can estimate $\eta$ as $10.8/11 = 0.98$.

\section{Conclusion}
In conclusion, this paper presents two key advancements in Large Language Model annotation (LLaMA) for punctuation restoration tasks. Firstly, the successful application of LLaMA for punctuation restoration demonstrates superior performance compared to the established Recurrent Neural Network Transducer (RNNT) model. Secondly, the introduction of Forward Pass Only Decoding (FPOD), a novel decoding approach that significantly improves inference speed. The experimental results validate the effectiveness of these methods, showing significant improvements in both the quality of punctuation restoration and inference speed. These findings open up new possibilities for future research and development in the field of punctuation restoration and natural language processing.
\bibliographystyle{IEEEtran}
\bibliography{mybib}

\end{document}